\documentclass{article}
\usepackage{kr}

\usepackage{times}
\usepackage{soul}
\usepackage{url}
\usepackage[hidelinks]{hyperref}
\usepackage[utf8]{inputenc}
\usepackage[small]{caption}
\usepackage{graphicx}
\usepackage{amsmath,amssymb,latexsym}
\usepackage{amsthm}
\usepackage{booktabs}
\usepackage[ruled,vlined]{algorithm2e}
\usepackage{xcolor}
\usepackage{color}
\definecolor{citecolor}{HTML}{000770}
\definecolor{urlcolor}{HTML}{0006b8} 
\hypersetup{pdfstartview=FitH,  citecolor=citecolor,urlcolor=urlcolor, colorlinks=true}

\usepackage{macros}

\newtheorem{definition}{Definition}

\title{Interpretable Reinforcement Learning with Multilevel Subgoal Discovery}

\author{%
Alexander Demin\and
Denis Ponomaryov \\
\affiliations
Ershov Institute of Informatics Systems, Novosibirsk, Russia \\
\emails
alexandredemin@yandex.ru,
ponom@iis.nsk.su
}

\begin{document}

\maketitle

\begin{abstract}
We propose a novel Reinforcement Learning model for discrete environments, which is inherently interpretable and supports the discovery of deep subgoal hierarchies. In the model, an agent learns information about environment in the form of  probabilistic rules, while policies for (sub)goals are learned as combinations thereof. No reward function is required for learning; an agent only needs to be given a primary goal to achieve. Subgoals of a goal $G$ from the hierarchy are computed as descriptions of states, which if previously achieved increase the total efficiency of the available policies for $G$. These state descriptions are introduced as new sensor predicates into the rule language of the agent, which allows for sensing important intermediate states and for updating environment rules and policies accordingly. 
\end{abstract}

\section{Introduction}
Hierarchical Reinforcement Learning (HRL) provides a solution to the problem of sample efficiency by the employment of hierarchical policy learning, which gives a clear advantage in comparison to ``flat'' RL models. The ability to discover subgoal hierarchies allows the agent for interacting with the environment in a more targeted way, which gives an increased learning speed and performance.

Most of the modern end-to-end HRL approaches 
support the discovery of subgoal hierarchies of a fixed depth. Only some of them do not have this limitation (see Table 2 in \cite{Pateria2021}), but they provide models, which are not inherently interpretable.

The development of the Deep Learning has greatly broadened the variety of environments and tasks approached by RL, but the problem of interpretabilty of the novel RL models became evident. Typically a subsymbolic RL model is augmented with semantic entities, which make the model behavior more transparent, or it is provided with a glass-box symbolic model, which approximates its behavior. In the recent years, quite a few novel RL approaches have been proposed, which are based on inherently interpretable models \cite{Puiutta2020}. 

Less attention in the modern RL is paid to tasks for discrete environments. However, many non-trivial tasks for RL originate in business and industry, where interpretable models supporting subgoal discovery are required. For example, a whole bunch of tasks is concerned with customer interaction and sales funnel operation, in which it is required to identify the key steps (subgoals) of interaction that lead customers to a purchase in order to guide them on relevant trajectories. Although data preparation and simulation for such environments are separate important problems, we believe that RL technologies for discrete state and action spaces should be further advanced and benchmarked in model environments.

In this paper, we propose a novel RL model for discrete environments, which is inherently interpretable and supports the discovery of arbitrary deep subgoal hierarchies. In our model, an agent learns information about environment in the form of probabilistic rules, while policies for (sub)goals are learned as combinations thereof. No reward function is required for learning; initially an agent only needs to be given a primary goal to achieve. Subgoals of a goal $G$ from the hierarchy are computed as descriptions of states, which, if previously achieved, increase the total efficiency of the available policies for $G$. These state descriptions are introduced as \emph{invented sensor predicates} into the rule language of the agent, which allows for sensing important intermediate states and for updating environment rules and policies accordingly. The model is implemented as a combination of environment rule, policy, and subgoal learning procedures, which are executed online in an interleaving fashion.

We evaluate our model on a Item Picking Task for grid world environments, in which items of types $1\lldots k$, for $k\geqslant 1$, are randomly distributed, with $k$ being a parameter of the environment. An agent navigating in the environment can pick up an item of each type only once. An item of type $i$, for $1\leqslant i\leqslant k$, can be picked up if either $i=1$, or the agent  has previously picked up some item of type $i-1$. The primary goal set to the agent is to pick up an item of type $k$. This setting is an abstraction of several prominent subgoal discovery tasks for grid world domains (e.g., the Taxi Domain, in which an agent must first pick up a passenger and then reach a certain destination, or the Multiple-Room Domain, in which an agent must pass a sequence of rooms/doorways in order to access a goal room) and it is general enough for testing the discovery of subgoals of various depths. 

In order to provide a ground for examples in the paper, we begin the exposition with a description of a particular instance of the Item Picking Task. Then in Section \ref{Sect:ModelArchitecture} we describe the general architecture of our RL model and consider its components in detail. We provide preliminary experimental results on the performance of the model in Section \ref{Sect:Experiments} and we compare our work with the most relevant approaches known in the literature in Section \ref{Sect:RelatedWork}. We conclude with a discussion of our approach and topics for future research in Section \ref{Sect:Outlook}.

\section{Example Environment}\label{Sect:ExampleEnvironment}
Reinforcement Learning models for discrete environments are typically benchmarked on grid world domains, in which an agent must learn how to achieve a certain goal state. As an example environment in the paper we consider a 2-dimensional grid of a fixed size, on which items of types $1\lldots k$, for $k\geqslant 1$, are randomly distributed. Every time an agent has a choice of three possible actions: $turn\!-\!left, turn\!-\!right, move$. While moving in the environment the agent picks up items according to the linear order on their types: an item of type $i$, for $1\leqslant i\leqslant k$, is picked up iff the agent has reached the position of this item and it has not previously picked up an item of type $i$, and either $i=1$, or the agent has previously picked up an item of type $i-1$. The primary goal set to the agent is to pick up an item of type $k$. Thus, the item types $1\lldots k-1$ correspond to the (sub)goals that the agent must consequently achieve on the way to the primary goal. 

\begin{figure}
\centering
\includegraphics[height=1.75in,keepaspectratio]{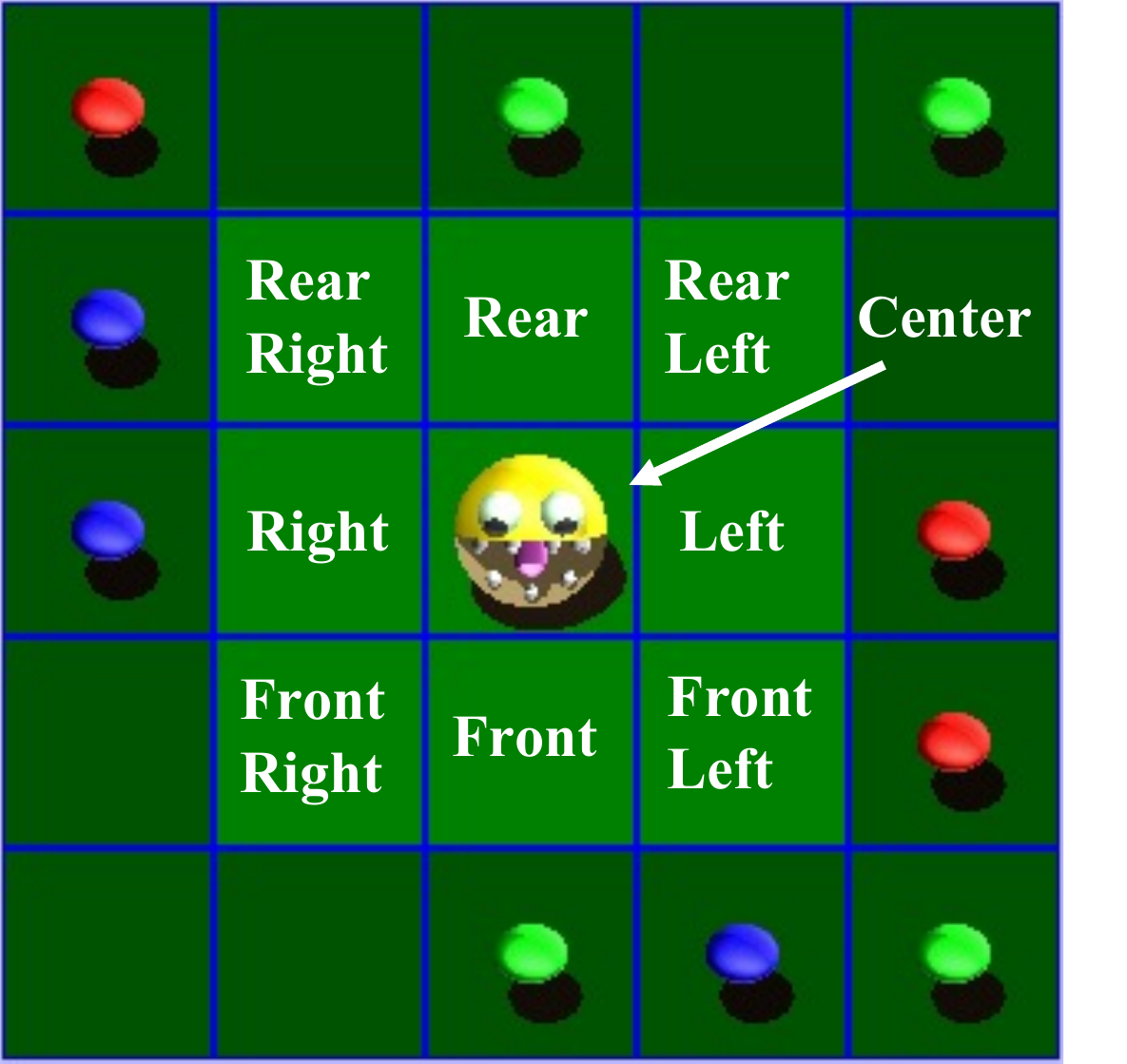}
\caption{Agent's sensor field}
\label{Fig:SensorField}
\end{figure}

The agent has ten sensors in total. Nine of them inform about the types of the items located in the cells near the agent (e.g., $Right(type3)$, see Figure \ref{Fig:SensorField}), or they indicate $empty$, if the corresponding cells are not occupied by any item (e.g., $Right(empty)$), or $wall$, respectively, if there is a grid boundary.
One more sensor named $PickedUp$ indicates whether the agent has picked up some item at current position. Thus, the agent is able to sense the fact of picking up something, but it cannot detect the type of an item it has just picked up, which in general makes the Item Picking Task more complex for the agent.


\section{Model Architecture}\label{Sect:ModelArchitecture}
\begin{figure}[h]
\centering
\makebox{
\put(-120,0){\includegraphics[height=2.3in,keepaspectratio]{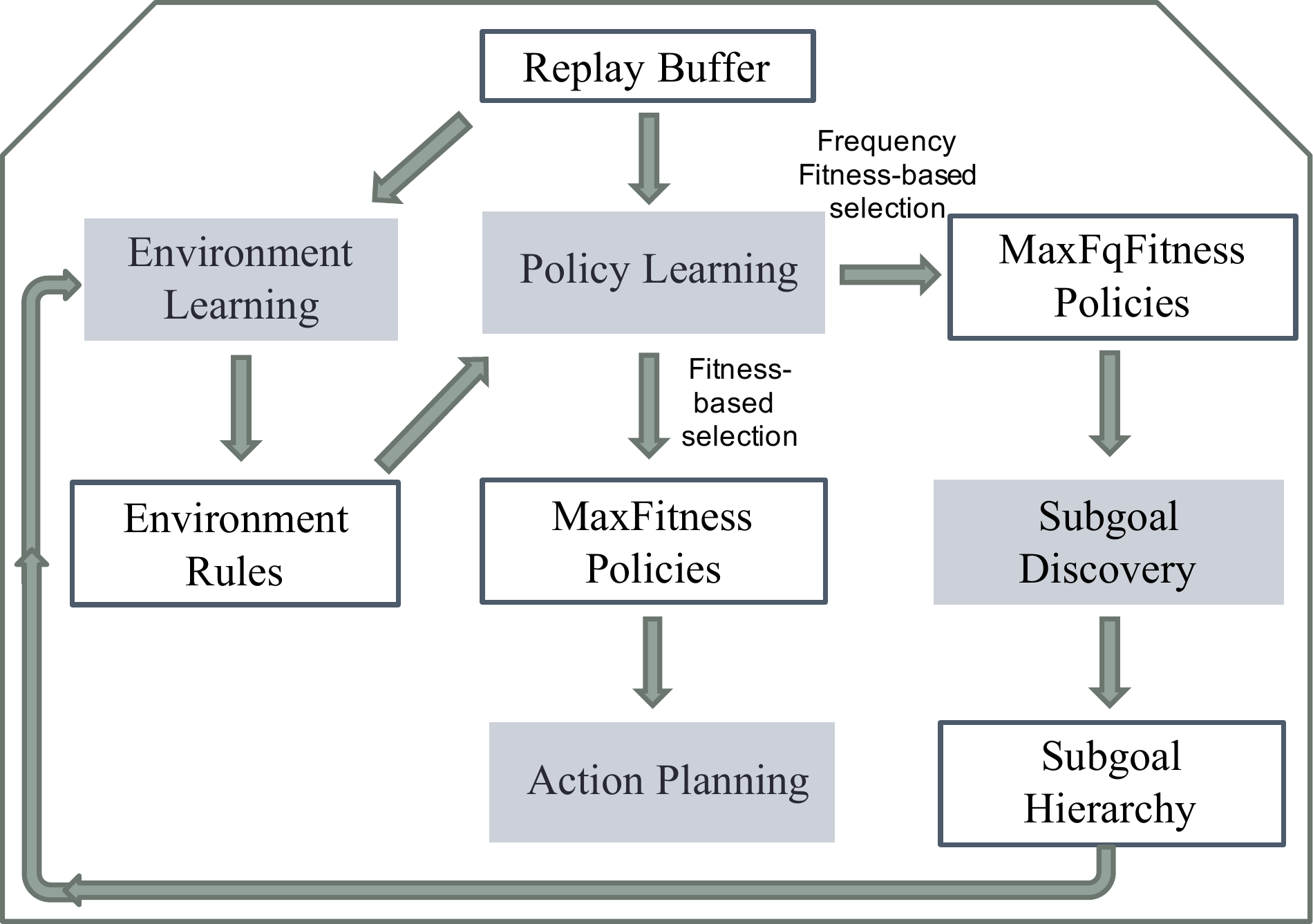}}
}
\caption{Data flow flow between modules of the model} 
\label{Fig:Dataflow}
\end{figure}

Our model is based on a combination of four modules, which are depicted on Figure \ref{Fig:Dataflow}. Initially, an agent is given a primary goal as a state description in terms of sensor predicates. For example, $Center(type3),  PickedUp$ describes states, in which the agent has reached an picked up an item of type $3$. After initialization, the agent performs a number $N$ of random actions in the environment. State transitions made by the agent are recorded in a replay buffer. Every round of $N$ actions is followed by the activation of the Environment Learning module, which learns the effects of agent's actions in the form of (probabilistic) rules. For instance, a typical rule for our example environment could look like 
{\footnotesize
\begin{equation}
Right(type1), turn\!-\!right \rightarrow Front(type1)
\end{equation}}
which means that turning right in the situation when an item of type $1$ is located to the right of the agent results in the situation when the item is in front of it. 

Environment Learning is followed by the activation of the Policy Learning module, which learns policies for all (sub)goal states as (probabilistic) rule-like expressions that represent state-action-state trajectories leading to (sub)goals. Policies and environment rules are ranked by the estimation of their probability on the replay buffer. For example, one of the highly ranked policies for a goal state $Center(type3),  PickedUp$ in our example environment would look as 
{\footnotesize \begin{equation}\label{Policy:Updated}
\begin{aligned}
R&ight(type3),  HasType2\  \{turn\!-\!right\} \\
& Front(type3), HasType2 \ \{move\} \\
& \ Center(type3), PickedUp
\end{aligned}
\end{equation}}
where $HasType2$ is a sensor predicate, which is true whenever the agent has previously picked up an item of type $2$. These auxiliary predicates are ``invented'' by the Subgoal Discovery Module launched after each round of $M$ actions (where $M\geqslant N$). For example, $HasType2$ would indicate that the agent has previously achieved the (subgoal) state $Center(type2), PickedUp$. 
Note that a variant of the policy above without this predicate would look as 
{\footnotesize \begin{equation}\label{Policy:Example}
\begin{aligned}
R&ight(type3),  \{turn\!-\!right\} \ Front(type3) \ \{move\} \\
& \ Center(type3), PickedUp
\end{aligned}
\end{equation}}
Intuitively, this policy would have a low rank, since the agent is not able to pick up an item of type $3$ without having previously picked up one of type $2$. Thus, many instances of the trajectory $Right(type3) \{turn\!-\!right\} Front(type3) \{move\} $ would not result in the state $Center(type3), PickedUp$. The idea behind the subgoal discovery in our model is that updating policies with predicates for important intermediate states should give policies with an increased rank on average. Subgoal Discovery is followed by the activation of the above mentioned learning modules, which recompute environmental rules and policies in the language with the invented predicates.

\begin{figure}[h]
\centering
\includegraphics[height=2.9in,keepaspectratio]{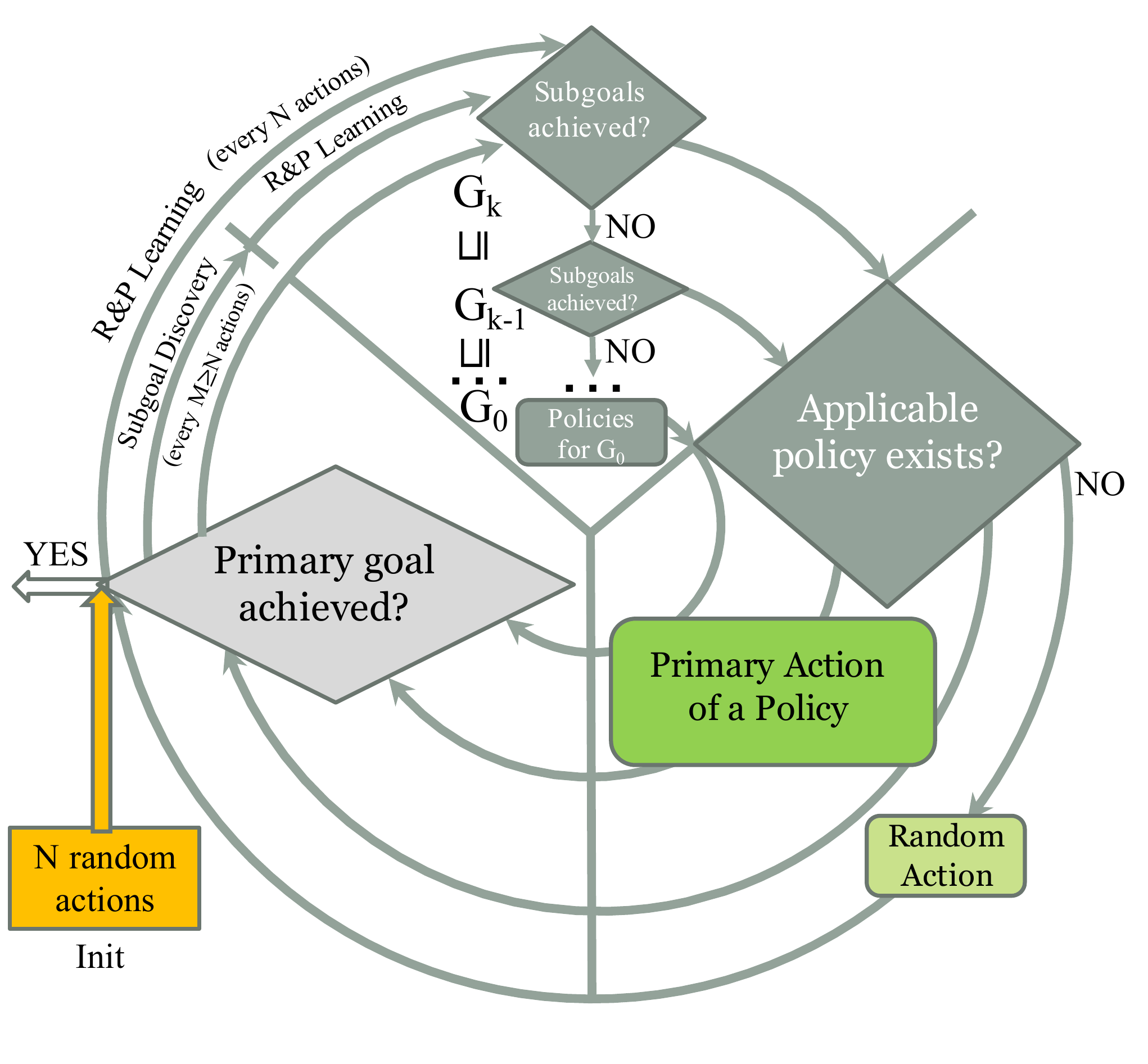}
\caption{Control flow between modules of the model. Here $G_0\!\sqsubseteq\! \ldots \!\sqsubseteq\! G_k$ is an example subgoal hierarchy and ``R\&P Learning'' stands for \emph{Environment Rule and Policy Learning}} 
\label{Fig:FlowChart}
\end{figure}

Finally, the Action Planning Module is used to decide on the next action depending on the policies applicable in the current situation and on the previously achieved subgoals. The module selects one of the highly ranked policies for the lowest non-achieved (sub)goal in the hierarchy, which is applicable in the current state (Figure \ref{Fig:FlowChart}). If there is no such policy, a random action is performed. Otherwise, the primary action from the policy is executed and the action planning process repeats. Thus, the agent does not have to make a complete sequence of actions from a single policy suitable in the current state; instead it can dynamically combine actions from different policies. 

In the next subsections we describe the components of our model in detail. 

\subsection{Environment Learning}\label{Sect:EnvironmentLearning}
In our model, an agent learns the effects of actions in the form of probabilistic rules given in the following language. The alphabet of the language is defined as the disjoint union of sets $\Sens\cup U_G\cup \Acts$, where $\Sens$ is a set of sensor predicate names, one for each combination of a sensor and a valid indication, $U_G$ is an infinite set of names for (sub)goal predicates, and $\Acts$ a set of action predicate names, one for each agent's possible action. Although the set $U_G$ is infinite, each time the agent has only a finite subset of (sub)goal predicates $\Goals\subseteq U_G$ at its disposal. Initially $\Goals$ is a singleton set, which consists of a predicate name $G_{prime}$, a notation for the primary goal. 

All predicates are nullary and the rule language is inherently propositional. However, in our examples we use shortcuts like $P(c)$ (e.g., $Right(type3)$), where $P$ is a sensor name and $c$ is one of valid indications of the sensor.  An agent's \emph{state} is a finite subset of $\Sens\cup\Goals$. We denote the primary goal state as $S_{prime}$. In the following, we identify a state $\{P_1\lldots P_n\}$, $n\geqslant 0$, with its syntactic representation as a (possibly empty) string $P_1\lldots P_n$ and for a state $S$, we denote by $|S|$ the number of predicates in $S$.  

\begin{definition}[Environment Rule]\label{De:EnvironmentalRule}
An \emph{environment} \emph{rule} is an expression of the form
\begin{equation}\label{Eq:EnvironmentRule}
S_1, A \rightarrow S_2
\end{equation} 
where $S_1, S_2$ are non-empty states and $A\in\Acts$. 
\end{definition}

For a rule $R$ of the form above we call $S_1$ and $S_2$ the \emph{premise} and \emph{conclusion} of $R$, respectively, and we use notations $\pre(R)$ and $\post(R)$. A \emph{refinement} of a rule $R$ is an environment rule $R'$ such that $\post(R')=\post(R)$, $\pre(R)\subset\pre(R')$, and $|\pre(R')|=|\pre(R)|+1$. 

Let $\langle \Omega, \preceq\rangle$ be a linearly ordered set. We slightly abuse the standard mathematical terminology an call a subset $\{\tau_1\lldots \tau_n\}$ of $\Omega$, where $n\geqslant 1$, a \emph{chain} if for all $i=1\lldots n-1$ it holds that $\tau_i\preceq\tau_{i+1}$ and for any $\tau\in\Omega$ if $\tau_i\preceq\tau\preceq\tau_{i+1}$ then $\tau=\tau_j$, for some $j\in\{i,i+1\}$.

\begin{definition}[Replay Buffer]\label{De:ReplayBuffer}
A \emph{replay buffer} is a triple $\langle \Omega, \sqsubseteq, \pi\rangle$, where $\sqsubseteq$ is a partial order on $\Goals$, with $G_{prime}$ being the supremum, $\pi:\Goals\mapsto 2^{\Sens}$ is a mapping such that $\pi(G_{prime})=S_{prime}$, and $\Omega$ is a finite linearly ordered (wrt a relation $\preceq$) set of tuples of the form $\langle S_{pre}, A, S_{post}\rangle$, where $A\in\Acts$ and $S_{pre} ,S_{post}$ are non-empty states, such that for any chain $\{\tau,\tau'\}$ of tuples $\tau=\langle S_{pre}, A, S_{post} \rangle $, $\tau'=\langle S'_{pre}, A', S'_{post} \rangle$ from $\Omega$  it holds that $S_{post}=S'_{pre}$.
\end{definition}
Intuitively, a replay buffer serves as a history of agent's transitions provided with the actual subgoal hierarchy $\sqsubseteq$ and an intepretation of subgoal predicates (in terms of states) via the mapping $\pi$. In the following, we omit references to the set $\Omega$ and say simply that a tuple $\tau$ is \emph{from} a replay buffer $\cal{B}$. We note that the notion of replay buffer could be reformulated in more strict mathematical terms as a many-sorted algebra.


A predicate $S\subseteq\Sens\cup\Acts$ is \emph{true on a tuple} $\tau=\langle S_{pre}, A, S_{post}\rangle$ from a replay buffer $\cal{B}$ if $S\in S_{pre}\cup\{A\}$. A subgoal predicate $G\in\Goals$ is true on $\tau$ if there is a tuple $\tau^{\uparrow}=\langle S^{\uparrow}_{pre}, A^{\uparrow}, S^{\uparrow}_{post}\rangle$ from $\cal{B}$ such that
\begin{itemize}
\item  $\tau^{\uparrow}\preceq\tau$, $\tau^{\uparrow}\neq\tau$
\item $\pi(G)\subseteq S^{\uparrow}_{post}$
\item for any tuple $\tau^{\downarrow}=\langle S^{\downarrow}_{pre}, A^{\downarrow}, S^{\downarrow}_{post}\rangle$ and subgoal $G'$, if $\tau^{\uparrow}\preceq\tau^{\downarrow}\preceq\tau$ and $G\sqsubset G'$ then  $\pi(G')\not\subseteq S^{\downarrow}_{post}$
\end{itemize}
Informally, the last condition means that the state corresponding to $G$ is achieved in some previous situation $\tau^{\uparrow}\prec\tau$ in the history and the achievement of subgoals is reset by higher goals.

A set $S\subseteq\Sens\cup\Goals\cup\Acts$ \emph{holds on a tuple}  $\tau$ from $\cal{B}$ (in symbols  $\tau\models S$) if every predicate from $S$ is true on $\tau$.
A rule $R=S_1, \ A \rightarrow S_2$ \emph{holds on a replay buffer} $\cal{B}$ with probability $p$ if  $prmconc / prm=p$, where  $prm$ is the number of tuples from $\cal{B}$, on which $S_1\cup\{A\}$ holds, and $prmconc$ is the number of tuples $\tau=\langle S_{pre}, A, S_{post}\rangle$ from $\cal{B}$ such that $S_1\cup\{A\}$ holds on $\tau$, $S_2\cap\Sens\subseteq S_{post}$, and $\pi(G)\subseteq S_{post}$, for all $G\in S_2\cap\Goals$. Intuitively, $p$ is the frequency probability of the event $\post(R)$ conditioned on the event $\pre(R)$ in the history given by $\cal{B}$ provided that every subgoal predicate $G$ from the conclusion of $R$ is viewed as the event of the achievement of $\pi(G)$. In the paper, we use the \emph{partial} function $\prob$, which for a rule $R$ and a replay buffer $\cal{B}$ gives the probability of $R$ on $\cal{B}$ as above, provided $prm\neq 0$.

\begin{definition}[Probabilistic Law]\label{De:ProbLaw}
A rule $R$ is a \emph{probabilistic law} wrt a replay buffer $\cal{B}$ if $\prob(R, {\cal{B}})$ is defined and $\prob(R', {\cal{B}})<\prob(R,{\cal{B}})$, for any rule $R'$ such that $\pre(R')\subset\pre(R)$.
\end{definition}

For the example environment from Section \ref{Sect:ExampleEnvironment}, the rule $R=Right(type3), turn\!-\!right \rightarrow Front(type3)$ has probability $1$ on any replay buffer $\cal{B}$, whenever $\prob(R, {\cal{B}})$ is defined. The value of $\prob(R,{\cal{B}})$ is undefined in case the agent did not experience the state $Right(type3)$, or did not make a turn to the right from this state. Otherwise $\prob(R,{\cal{B}})=1$, because the environment is non-stochastic and thus, turning right in the situation, when an item is located to the right of the agent always brings it to the situation, in which the item is located in front of it. If $\prob(R,{\cal{B}})=1$ then it holds that $R$ is a probabilistic law. 

By definition, probabilistic laws with a probability value $p$ are exactly those rules, which by inclusion of premises, are the shortest ones among all the rules having a probability at least $p$. Thus, the concept of a probabilistic law in our model captures the balance between the size and ``informativeness'' of environment descriptions, the topic discussed broadly in AI. It is known that searching for the shortest implications true (with probability $1$) on a given data is a computationally hard problem. There may exist exponentially many such shortest implications (Theorem 1 in \cite{Kuznetsov2004}) and the problem to decide whether there is one of a given size is NP-complete \cite{Gunopulos2003}.

The environment learning procedure in our model is implemented as a heuristic-based enumeration of rules by refinement, with the selection of those ones, which satisfy the properties of a probabilistic law. The procedure is given by Algorithm \ref{Algo:EnvironmentLearning}. 

\begin{algorithm}\caption{Environment Rule Learning}\label{Algo:EnvironmentLearning}

  \footnotesize
  \DontPrintSemicolon

  \SetKwInOut{Input}{Input}
  \SetKwInOut{Output}{Output}
  \SetKwInOut{Params}{Parameter}
  \SetKwComment{Comment}{}{}
  
  \Input{Replay buffer $\cal{B}$,  a non-empty state $S$}
  \Params{Base enumeration depth $d\geqslant 1$}
  \Output{A set of prob. laws wrt $\cal{B}$ with conclusion $S$}

  \Begin{
    \nl RULES := $\{R \mid R \ \text{is a rule, with} \ \post(R)=S, \ |\pre(R)|\leqslant d\}$     \tcc*[f]{environment rule enumeration}  \;
    \nl LWS := $\{R\in \text{RULES} \mid R \ \text{is a probabilistic law} \}$\;
    \nl $2Refine$ := $\{R\in \text{LWS} \mid \neg\exists R'\in\text{LWS} \ \text{s.t.} \ \pre(R)\subset\pre(R') \}$\;
    \nl\While{$2Refine\neq\varnothing$} {
    	\nl Let $R\!\in\!2Refine$; $\ \ $ $2Refine:=2Refine\!\setminus\!\{R\}$\;
       	\nl\ForEach{refinement $R'$ of $R$}{
    	  \If {$R'$ is a probabilistic law}{
			\nl $2Refine:=2Refine\cup\{R'\}$\;
			\nl LWS := LWS$\cup\{R'\}$\;
		  }
    	}
    }    
    
   \nl \textbf{return} LWS\;
 }   
\end{algorithm}


The heuristic in the algorithm implies that after base enumeration up to depth $d$ (which is one of the hyperparameters of our model) a rule is selected as an additional candidate member for the resulting set of laws only if it refines one of the probabilistic laws $R$ obtained by base enumeration and has a strictly higher probability than $R$. Monotonicity in this sense does not hold in general. It can be the case that every refinement of a rule $R$ has a lower probability than $R$, but there is a rule $R'$ obtained by extending the premise of $R$, e.g., with two predicates, such that $R'$ is a probabilistic law. Our algorithm can find the law $R'$ only if the parameter $d$ is greater than $|\pre(R')|$. 

The assumption behind the heuristic is that state descriptions, which are most informative for learning the environment, usually involve a moderate number of sensor predicates. Hence, the majority of probabilistic laws could be found by the algorithm with a small parameter $d$. By relying further on monotonicity, the algorithm tries to find probabilistic laws with higher probabilities. This is implemented by computing refinements of probabilistic laws obtained by base enumeration incrementally as long as this gives new laws. 

Besides $d$, an implementation of Algorithm \ref{Algo:EnvironmentLearning} employs the following hyperparameters for fine tuning. To exclude laws from the output, which have a low probability or statistical significance (on the replay buffer), the parameters \emph{Probability-Threshold} and \emph{Confidence-Threshold} are used, respectively. \emph{Probability-Gain-Threshold} is used to exclude those laws from further refinement, for which the last refinement step has given a low probability gain. Finally, the \emph{Max-Sensor-Predicates} parameter restricts the search to the rules with a given maximal number of sensor predicates in the premise. We comment on the hyperparameter settings for experiments in Section \ref{Sect:Experiments}.

\subsection{Policy Learning}\label{Sect:PolicyLearning}

\begin{definition}[Policy]\label{De:Policy}
A \emph{policy} $P$ for a non-empty state $G$ is an expression of the form
\begin{equation}\label{Eq:PolicyRule}
S_1\ \{A_1\} \ldots S_n \ \{A_n\} \ G
\end{equation} 
where $n\geqslant 1$ and for all $i=1\lldots n$, $A_i\in\Acts$ is an action predicate and $S_i$ a non-empty state such that $S_i\cap\Goals=S$, for a (possibly empty) set $S$.
\end{definition}

Intuitively, the policy gives a state-action-state trajectory leading to state $G$. The last condition in the definition implies that subgoals are never lost along the way to $G$ and they cannot be suddenly achieved at some point of the trajectory. Thus, the policy describes only how the target state $G$ could be achieved, while the ways of how all the required subgoals could be achieved are to be given by separate policies.

For a policy $P$ of the form above the set $S_1$ is called the \emph{policy premise} (we abuse notation and write $\pre(P)$ to denote the premise of $P$) and $A_1$ is called the \emph{primary action} of $P$.  The number $n$ is called the \emph{length} of the policy and it is denoted as $len(P)$. 

\begin{definition}[Policy Fitness]\label{De:Fitness}
Given a replay buffer $\cal{B}$, the \emph{fitness} of a policy $P=S_1\ \{A_1\} \ldots S_n \ \{A_n\} \ S_{n+1}$ wrt $\cal{B}$ (in symbols $\fitness(P,{\cal{B}})$) is the product of probabilities $\prob(S_i, A \rightarrow S_{i+1}, \ \cal{B})$, for all $i=1\lldots n$, if each of them is defined. Otherwise the fitness of $P$ is undefined. 
\end{definition}

Let $P=S_1\ \{A_1\} \ldots S_n \ \{A_n\} \ G$ be a policy and $R=S_0, A_0\rightarrow S_1$ an environment rule such that 
$S_0\cap\Goals=S_1\cap\Goals$. A \emph{refinement} of $P$ 
with $R$ denoted as $\REFN(P,R)$ is the policy $S_0 \ \{A_0\} \ S_1\ \{A_1\} \ldots S_n \ \{A_n\} \ G$. If $P$ and $P'$ are policies for the same state then $P'$ is called a \emph{variant} of $P$ if $\pre(P')\subseteq\pre(P)$ and $P\neq P'$. That is, a variant policy provides an alternative trajectory to achieve the same target state from the same or a more general situation. 


For a replay buffer $\cal{B}$ and a state $S$, let \Algo{\ref{Algo:EnvironmentLearning}}$({\cal{B}},S)$ denote the set of probabilistic laws obtained by Algorithm \ref{Algo:EnvironmentLearning} on the input\footnote{we assume that all hyperparameters for algorithms are fixed globally.} ${\cal{B}}, S$. We now describe a procedure for computing a set of policies for a given state $G$. The procedure builds every policy expression backwards, starting from $G$, by incremental refinement with probabilistic laws computed by Algorithm \ref{Algo:EnvironmentLearning}. 
A policy obtained this way may thus represent a trajectory that has not been seen as a whole by the agent before. Only particular transitions (given by probabilistic laws) could have been experienced by the agent. This enables the agent to ``reason'' about unseen trajectories, which contributes to the sample efficiency of our model. 

The policy learning algorithm is given below. By the definition of the fitness function, longer policies get lower fitness values, therefore the algorithm computes \emph{strong} policies, i.e. those, which have the greatest fitness value among all variants. 

\begin{algorithm}\caption{Policy Learning}\label{Algo:PolicyLearning}

  \footnotesize
  \DontPrintSemicolon

  \SetKwInOut{Input}{Input}
  \SetKwInOut{Output}{Output}
  \SetKwInOut{Params}{Parameter}
  \SetKwComment{Comment}{}{}
  \SetKwProg{Fn}{Function}{}{end function}

\Input{Replay buffer $\cal{B}$, non-empty state $G$}
\Output{A set of policies for $G$}
  
  \Begin{

\nl LWS := \Algo{\ref{Algo:EnvironmentLearning}}$({\cal{B}}, G)$\;
\nl Let POL := $\{S_{pre} \ \{A\} \ G \ \mid \ S_{pre}, A \rightarrow G\in$ LWS $\}$ \tcc*[f]{initialize a set of policies}\;

\BlankLine
\nl Let $2Process$ := POL; $\ \ $  RPol := $\varnothing$\;
\BlankLine
\nl\While {$2Process\neq\varnothing$}{
	\nl \For {$policy\in 2Process$} {
		\nl LWS := filterbyGoals(\Algo{\ref{Algo:EnvironmentLearning}}$({\cal{B}},\pre(policy))$)\;
		\nl RPol := RPol $\cup$ $\{\REFN(policy, R) \mid R\in \ \text{LWS}\}$\;
	}
    \nl $2Process$ := getStrong(RPol, POL, $\cal{B}$)\;
	\nl POL := POL $\cup \ 2Process$ \;
}
\nl \textbf{return} POL\;



  

  }
\end{algorithm}

\begin{function}\caption{filterbyGoals($rules$)}\label{Algo:FilterByGoals}
  \footnotesize
  \DontPrintSemicolon

\nl\ForAll {$R\in$rules}{
			\nl\If {$\pre(R)\cap\Goals\neq \post(R)\cap\Goals$}{
			\nl rules := rules$\setminus\{R\}$\;
			}
}
\nl \textbf{return} $rules$\;
\end{function}

\begin{function}\caption{getStrong($policies,wrtpolicies,rBuffer$)}\label{Algo:getStrong}
  \footnotesize
  \DontPrintSemicolon
  
  \nl\ForEach {$P\in policies$}{
		\nl\If {$\exists$ variant $P'\in wrtpolicies$ of $P$ s.t.  $\fitness(P, \text{rBuffer})\leqslant\fitness(P',\text{rBuffer})$ and $len(P')<len(P)$}{
		\nl $policies:=policies\setminus\{P\}$\;
		}
}
\nl \textbf{return} policies
\end{function}
An implementation of Algorithm \ref{Algo:PolicyLearning} employs two hyperparameters for fine-tuning: the \emph{Maximal-Policy-Length} parameter is used to restrict the search only to policies of a fixed maximal length and \emph{Fitness-Gain-Threshold} prevents further refinement of those policies, whose fitness is below a given value. 

\subsection{Subgoal Discovery}
Recall policy (\ref{Policy:Example}) for our example environment,
which is composed of probabilistic laws $R_1=Right(type3), \ turn\!-\!right \rightarrow Front(type3)$ and  $R_2=Front(type3), \ move \rightarrow Center(type3), PickedUp$. Let us denote this policy as $P$. Even if the rule $R_1$ has probability $1$ (on a replay buffer $\cal{B}$), the fitness of $P$ may be quite low if so is the probability of $R_2$ (i.e., if only in few situations in the history the agent was able to pick up an item of type $3$ in front of it). However, if we consider only those situations, when it is known that the agent has previously picked up an item of type $2$ 
then the picture becomes different. If an item of type $3$ happens to be in front of the agent in one of these situations and the action $move$ is made, then with high probability the agent picks up this item. 

If the alphabet $\Goals$ is extended with a predicate named, e.g., $HasType2$, which is true whenever the agent has previously achieved the state $Center(type2), PickedUp$, then we can consider an update of the rule $R_2$ as $R'_2=Front(type3), HasType2, move \rightarrow Center(type3), PickedUp$, which is now (by an argument as in Section \ref{Sect:EnvironmentLearning}) a candidate for being a probabilistic law. 

Similarly, we can consider an update of $R_1$ as $R'_1=Right(type3), HasType2, \ turn\!-\!right \rightarrow Front(type3), HasType2$ obtained by adding the ``invented'' predicate to the premise and conclusion of $R_1$. 

Finally, consider an update $P'$ of the policy $P$ as (\ref{Policy:Updated}) in Section \ref{Sect:ModelArchitecture}.
If the probability of the both rules $R'_1$ and $R'_2$ is $1$, then so is the fitness of $P$, i.e., $P$ gives the most plausible trajectory to achieve the goal state $G=\{Center(type3),$ $PickedUp\}$ from the state $\{Right(type3), HasType2\}$, where $HasType2$ is a predicate for the \emph{subgoal} state $\{Center(type2), PickedUp\}$ of $G$. 

The idea behind the subgoal discovery in our model is that updating policies for a goal $G$ with the ``invented'' subgoal predicates as above should provide policies, which more probably lead to $G$, than the original ones. Given a replay buffer $\cal{B}$, the subgoal states for a state $G$ are searched as  combinations of sensor predicates, which are subsets of states visited by the agent. A discovered subgoal state $S$ gets a fresh predicate name ${\tt P}$ from $U_G\setminus\Goals$ as a shortcut, the set $\Goals$ is extended with ${\tt P}$, and the value of the interpretation function $\pi$ for ${\tt P}$ is defined as $S$. 

Let us note however that the policy $P$ (and hence, $P'$ in our example) may represent a trajectory that the agent has never experienced, i.e., there may exist separate tuples $\tau=\langle \{Right(type3),HasType2\}, \ turn\!-\!right, \ \{Front(type3),HasType2\}\rangle$ and $\tau'=\langle \{Front(type3),HasType2\}, \ move, \  \{Center(type3),$ $PickedUp\} \rangle$ in the replay buffer, but no chain $\{\tau, \tau'\}$. Therefore, to estimate the potential increase of the efficiency of policies after an update with a candidate subgoal, we take into account only those policies, which correspond to trajectories passed by the agent. For this, we introduce a frequency probability measure, called \emph{frequency fitness}, which indicates for a policy $S_1 \{A_1\} \ldots S_{n}, \{A_{n}\} \ G$, $n\geqslant 1$,  how often the agent achieved the goal state $G$ by following the trajectory $S_1 \{A_1\}  \ldots S_{n}, \{A_{n}\}$.

Let $\cal{B}$ be a replay buffer and $P=S_1\ \{A_1\} \ldots S_n \ \{A_n\} \ G$ a policy,  where $n\geqslant 1$. A tuple $\tau_{1}$ is called a (possible) \emph{starting point} of $P$ in $\cal{B}$ if there is a chain of tuples $\{\tau_1,.., \tau_{n}\}$ from $\cal{B}$ such that $\tau_i\models S_i\cup\{A_i\}$, for all $i=1,..,n$. A tuple $\tau_{n}=\langle S_{pre}, A_n, S_{post} \rangle$ from $\cal{B}$ is called a (possible) \emph{ending point} of $P$ in $\cal{B}$ if there is a chain $\{\tau_1,.., \tau_{n}\}$ in $\cal{B}$, which satisfies the above condition, where $\tau_n=\langle S_{pre}, A_n, S_{post} \rangle$ is a tuple such that $G\cap\Sens\subseteq S_{post}$ and  $\pi(S)\subseteq S_{post}$, for any $S\in G\cap\Goals$.

\begin{definition}[Frequency Fitness]\label{De:FrequencyFitness}
For a policy $P$ and a replay buffer $\cal{B}$, the \emph{frequency fitness} of $P$ wrt $\cal{B}$ (denoted as $\fqfitness(P,{\cal{B}})$) is defined as $E/S$ if $S\neq 0$, where $E$ and $S$ is the number of ending and starting points of $P$ in $\cal{B}$, respectively, and it is undefined otherwise.
\end{definition}

For a replay buffer $\cal{B}$ and a state $G$, let \Algo{\ref{Algo:PolicyLearning}}$({\cal{B}},G,\fqfitness)$ be the set of rules computed by Algorithm \ref{Algo:PolicyLearning}, which employs the frequency fitness instead of the original fitness measure. The procedure for the discovery of subgoals for a state $G$ is given by Algorithm \ref{Algo:SubgoalDiscovery}. 

For every tuple $\tau$ witnessing the achievement of $G$, the algorithm selects those policies from \Algo{\ref{Algo:PolicyLearning}}$({\cal{B}},G,\fqfitness)$ with the highest frequency fitness, for which $\tau$ is an ending point. Then the family of the best policies computed for all witness tuples is taken as an ``aggregate'' policy and its frequency fitness (function AvgFqFitness given below) is estimated before and after an update with a candidate subgoal state. Those states, wrt which the fitness gain is greater or equal a certain threshold $\beta$ (which is a hyperparameter), are taken as subgoals of $G$ and are introduced as fresh subgoal predicate names into the language of the agent.

\begin{algorithm}[h]\caption{Subgoal Discovery}\label{Algo:SubgoalDiscovery}

  \footnotesize
  \DontPrintSemicolon

  \SetKwInOut{Input}{Input}
  \SetKwInOut{Output}{Output}
  \SetKwInOut{Params}{Parameter}
  \SetKwComment{Comment}{}{}
  \SetKwProg{Fn}{Function}{}{end function}

\Input{Replay buffer $\cal{B}$ and $G\in\Goals$}
\Params{Base fitness gain $\beta$}
\Output{An update of $\cal{B}$ and $\Goals$}

\Begin{

\nl Let POL := \Algo{\ref{Algo:PolicyLearning}}$({\cal{B}},G,\fqfitness)$\;
\nl Let $BESTPOL:=\varnothing$\;

\nl\ForAll {$\tau=\langle S_{pre}, A, S_{post}\rangle$ from $\cal{B}$ s.t. $\pi(G)\subseteq S_{post}$} {
	\nl Let $P_\tau$ := $\{p\in\text{POL} \mid \tau \ \text{is an ending point of} \ p\}$\;
	\nl $BESTPOL\!:=\!BESTPOL \cup \{p\in P_\tau \mid \neg\exists p'\!\in\! P_\tau$ $\ \text{s.t.} \ \fqfitness(p,\!{\cal{B}})\!<\!\fqfitness(p',\!{\cal{B}})\}$\;
}

\nl\If {$BESTPOL = \varnothing$}{
	\nl \textbf{return} $\cal{B}$, $\Goals$  \tcc*[f]{no subgoal can be found}\;
	}


\nl\ForAll {$S\subseteq\Sens$ s.t. $S$ holds on a tuple from $\cal{B}$}{

	\nl\If { $\neg\exists P_S \in\Goals$ s.t. $\pi(P_S)=S$ and $P_S\sqsubseteq G$} { 
	\nl Let $P_S\in U_G\setminus\Goals$;  $\ \pi(P_S):=S$; $\Goals:=\Goals\cup\{P_S\}$ \tcc*[f]{invented predicate for S}\;

	\nl Let $UPDPOL:=\varnothing$\;
	\nl \ForAll {$S_1 \{A_1\} \ldots S_{n} \{A_{n}\} \ G\in BESTPOL$} {
		\nl $UPDPOL:=UPDPOL \ \cup$ $\{ S_1,P_S \ \{A_1\} \ldots S_{n}, P_S \ \{A_{n}\} \ G \}$\;
		}

\nl\If {AvgFqFitness(UPDPOL,$\cal{B}$)-AvgFqFitness(BESTPOL, $\cal{B}) \geqslant \beta$} {
	\nl  $\sqsubseteq:=\sqsubseteq\cup\{P_S,G\}$ \tcc*[f]{inserted $P_S$ into subgoal hierarchy}\;
	}
	\nl\Else{
	    \nl $\Goals := \Goals\setminus\{P_S\}$ \tcc*[f]{removed useless $P_S$}\;
	}
 }
}

\nl \textbf{return} $\cal{B}$, $\Goals$\;



 }
\end{algorithm}

\begin{function}[h!]\caption{AvgFqFitness($pol, rBuf$) }\label{Algo:AvgFqFitness}
\footnotesize
  \DontPrintSemicolon
  
\nl s := no. of $\tau$ s.t. $\exists p\in pol  (\tau \ \text{is a start. point of} \ p \ \text{in} \ rBuf )$\;
\nl e := no. of $\tau$ s.t. $\exists p\!\in\! pol (\tau \ \text{is an end. point of} \ p \ \text{in} \ rBuf )$\;
\nl \textbf{return} $0$ if $s=0$ or $\text{e}/\text{s}$, otherwise\;
\end{function}

\subsection{Action Planning}\label{Sect:ActionPlanning}
For a (subgoal) state $G$, let \Algo{\ref{Algo:PolicyLearning}}$({\cal{B}}, G)$ denote the set of policies for $G$ computed by Algorithm \ref{Algo:PolicyLearning}. The action planning algorithm given as a function below essentially makes ranking of the available policies for $G$ depending on the current state of the agent (the truth values of the sensor and subgoal predicates) and the ranking of policies for subgoals of $G$. The procedure either outputs a random action (in case there is no policy applicable in the current state), or the primary action of the best policy for $G$ if all subgoals of $G$ are achieved, or the primary action of a policy for a minimal (wrt $\sqsubseteq$) non-achieved subgoal of $G$. 
\begin{function}[h]\caption{Action Planning(\textit{(sub)goal state} $G$, \textit{replay buffer} $\cal{B}$)}\label{Algo:ActionPlanning}

  \footnotesize
  \DontPrintSemicolon



\nl Let $isRandom$ := true \tcc*[f]{random action flag}\;


\nl Let $curstate$ be the maximal tuple wrt $\preceq$ in ${\cal{B}}$  \tcc*[f]{current situation}\;
\nl Let $BESTPOL := getBestPol(G, curstate, \cal{B}$)\; 
\BlankLine
\nl\While {$BESTPOL\neq\varnothing$} {
	\nl Let $P \in BESTPOL$  \tcc*[f]{non-deterministic choice}\;
	\nl $BESTPOL := BESTPOL \setminus \{P\}$\; 
	\nl \If{$Rank(P, curstate, {\cal{B}})\neq 0$}{ 
			\nl $subG := \pre(P)\cap\Goals$ \tcc*[f]{subgoals of $G$ in policy $P$}\;
			\nl\If{$curstate\models subG$} { 
				\nl \textbf{return} $\langle$ primary action of $P$, $\neg isRandom \rangle$\; 
				}

			\nl Let $S\in subG$ s.t. $curstate\not\models S$ \tcc*[f]{non-deterministic choice}\;
			\nl Let $\langle a, israndm\rangle$ := Action Planning(S, ${\cal{B}}$)\;
			\nl\If {$\neg israndm$}{
				\nl \textbf{return} $\langle a,  false\rangle$\;
			}
}
}
\nl \textbf{return} $\langle$random action from $\Acts$, $isRandom \rangle$ \tcc*[f]{no policy applicable in the curr.state}\;
\end{function}
\begin{function}[h]\caption{getBestPol($S$, $curstate$, $\cal{B}$)}

  \footnotesize
  \DontPrintSemicolon

\nl Let POL := \Algo{\ref{Algo:PolicyLearning}}$({\cal{B}}, S)$\;
\nl \textbf{return} $\{P\in \text{POL} \ \mid \ \forall P'\in \text{POL}\ $  $Rank(P, curstate, {\cal{B}})\geqslant$ $Rank(P', curstate, {\cal{B}})\}$\; 
\end{function}
\begin{function}[h]\caption{Rank($policy$, $curstate$, $\cal{B}$)}

  \footnotesize
  \DontPrintSemicolon
  
\nl\If{curstate $\not\models\pre(policy)\setminus\Goals$} { 
\nl \textbf{return} 0 \tcc*[f]{policy not applicable in the current situation}\;
}

\nl Let subG := $\pre(policy)\cap\Goals$\;
\nl\If{$curstate\models$ subG} { 
\nl \textbf{return} $\fitness(policy, \cal{B})$  \tcc*[f]{all subgoals from $policy$ (if any) are achieved}\;
}
 

\nl \ForAll {$S\in\text{subG}$} {
\nl Let P[$S$] $\in$ $getBestPol$($S$, $curstate$, $\cal{B}$)\;
}

\nl \textbf{return} $\fitness(policy, \cal{B})\times\prod_{S\in\text{subG}}$Rank(P[$S$], $curstate$, $\cal{B}$)\;

\end{function}

\section{Preliminary Experimental Results}\label{Sect:Experiments}
For experiments we used an implementation of the environment for the Item Picking task introduced in Section \ref{Sect:ExampleEnvironment} on a grid of dimension 25x25. The environment was provided with the feature that each time an agent picked up some item an item of the same type appeared at a random non-occupied position on the grid. To support continuous learning, the ``semantics'' of the subgoal predicates was modified in such a way that as soon as the agent achieved the primary goal, i.e., picked up an item of the maximal type $k$, all the subgoal predicates became false. The fact of the achievement of an item of the maximal type was recorded in a primary-goal-counter and then the agent was set to achieve the primary goal again. In the experiments, the agent performance was measured in environments with $k=1,2,3$ by the number of the primary goals achieved in a course of $1000$ actions. The hyperparameters of the algorithms were set as follows. For Algorithm \ref{Algo:EnvironmentLearning}: base enumeration depth d=$3$, Probability-Threshold=$0.1$, Confidence-Threshold=$0.9$, Probability-Gain-Threshold=$0.1$, Max-Sensor-Predicates=$1$. For Algorithm \ref{Algo:PolicyLearning}: Maximal-Policy-Length=$4$, Fitness-Gain-Threshold=$0.5$. For Algorithm \ref{Algo:SubgoalDiscovery}, the base fitness gain parameter $\beta$ was set to $0.2$.

\begin{figure}[h]
\centering
\includegraphics[width=3in]{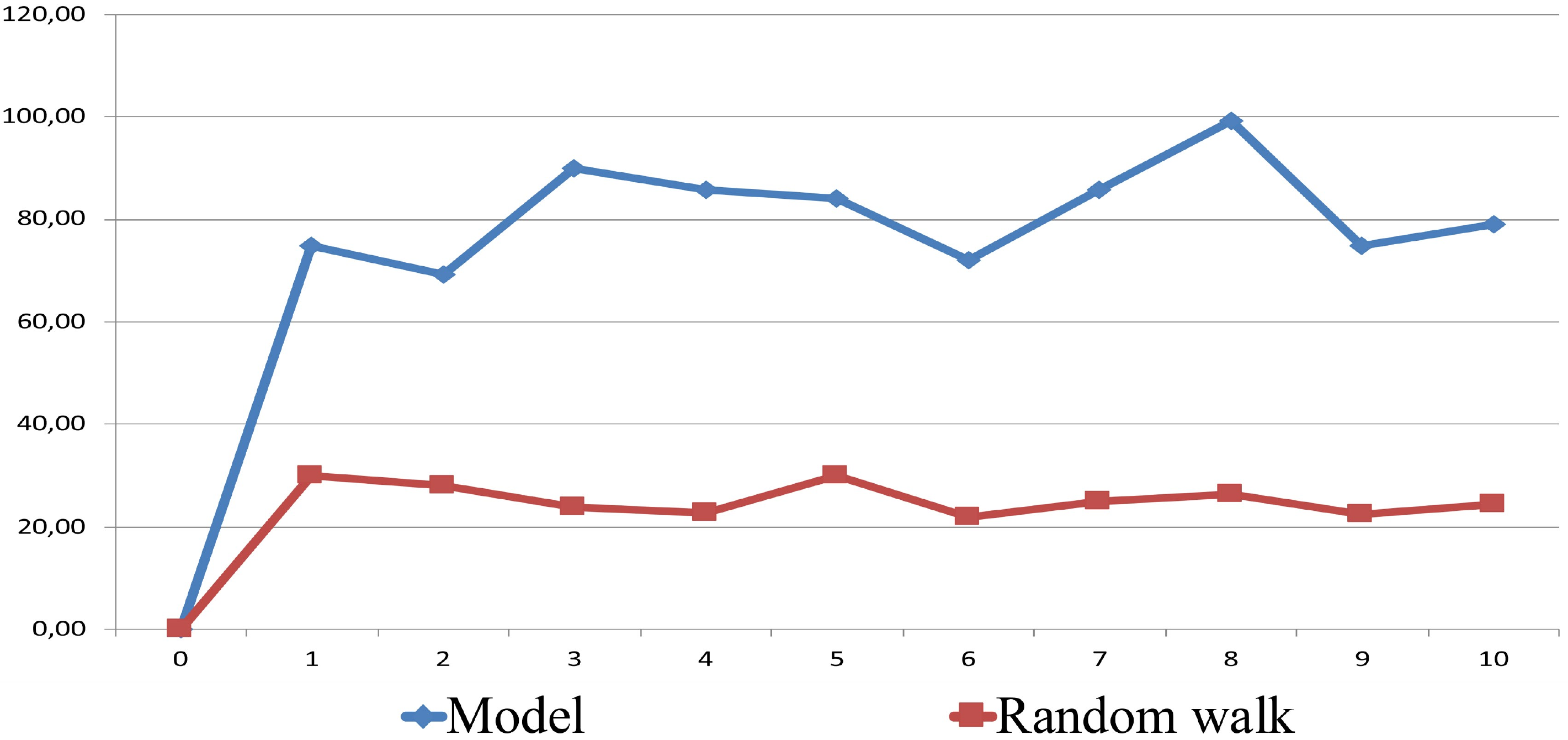}
\caption{Performance in environment for $k=1$}
\label{Fig:Exp1}
\end{figure}

In the experiment for $k=1$, the number $N$ of initial random actions was equal to $100$ and for $k=2,3$ it was set to $2000$. The upper bound on the number of actions was $10000$ in every experiment. Each of the modules of our model (See Figure \ref{Fig:Dataflow}) was initiated every $N$ steps. To smooth the effects of random item distribution in the environment we averaged the performance measurement on ten runs in each experiment. Figure \ref{Fig:Exp1} presents experimental results for the environment with $k=1$, in which the item type hierarchy is degenerate. 
In this experiment the agent did not discover any subgoal and it showed a relatively high performance already after the first $100$ steps. As subgoals were not used, the achieved performance was only due to the work of the Environment Learning and Policy Learning modules. 

\begin{figure}[h]
\centering
\includegraphics[width=3in]{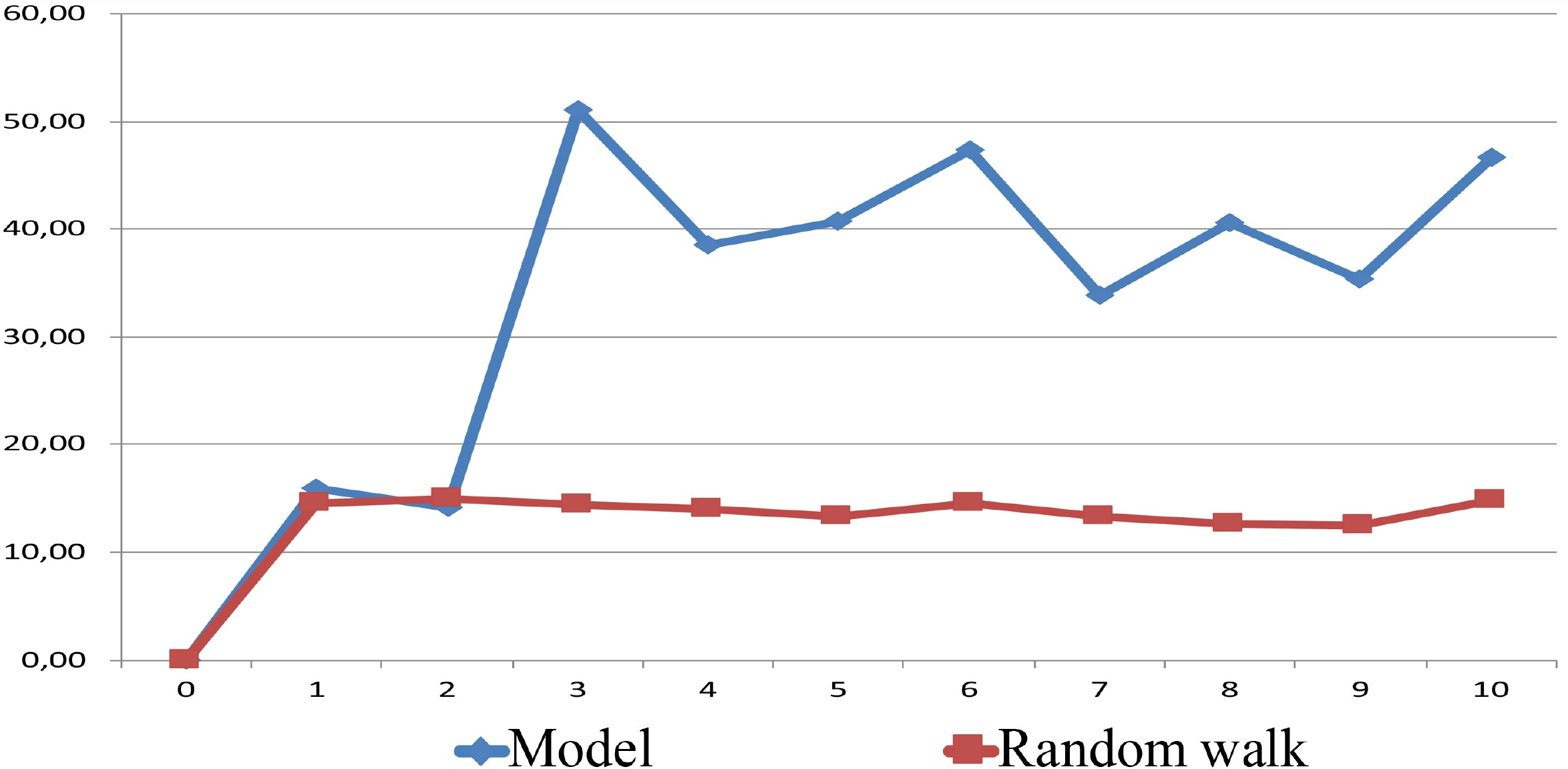}
\caption{Performance in environment for $k=2$}
\label{Fig:Exp2}
\end{figure}

\begin{figure}[h]
\centering
\includegraphics[width=3in]{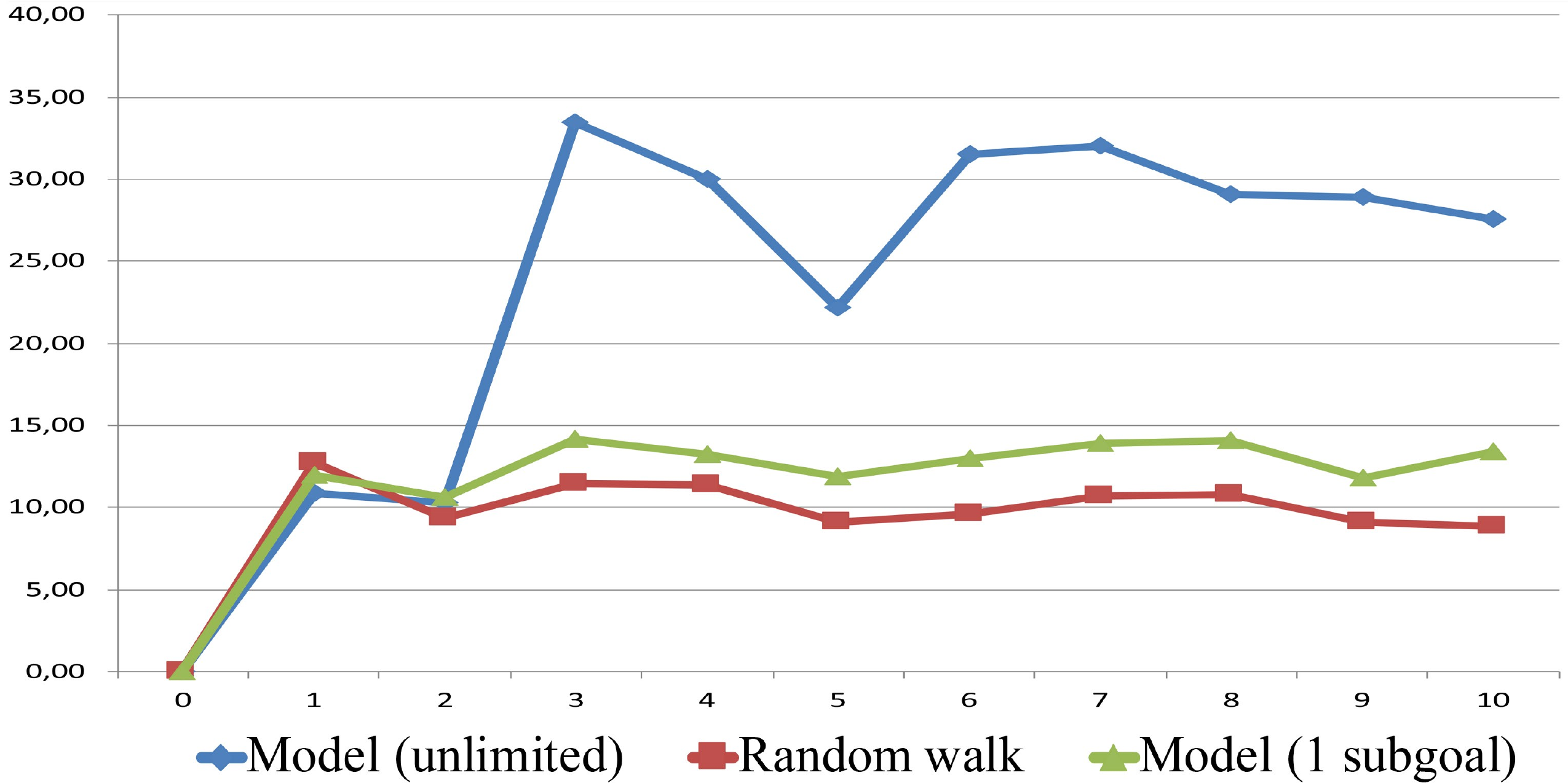}
\caption{Performance in environment for $k=3$}
\label{Fig:Exp3}
\end{figure}

The performance rate was non-stable in experiments, which is explained by the following fact. The sensor field of the agent is local and  allows for acting efficiently only in the proximity of an item. Our model does not solve the problem of efficient exploration of the environment when the items are out of reach of agent's sensors (this is one of the topics for future research). As a consequence, if an agent cleans up a certain region by picking up items, while new items are randomly generated distantly elsewhere, then it has to make quite a number of (random) steps to reach them, which yields a decreased performance. The performance change is even more radical in the environments with several item types (Figures \ref{Fig:Exp2}, \ref{Fig:Exp3}), since items of a required type may happen to be located even more distantly on average from the current agent's position than in the environment with a single item type. In each of the experiments the local performance maxima correspond to the ``islands'' in the environment, in which there was enough items of required types and thus, the agent was able to use its sensor field efficiently.

For the environment with $k=3$, Figure \ref{Fig:Exp3} shows the relative performance of an agent with unlimited subgoal capacity to that of an agent, which is able to discover a single subgoal. 
The result shows that the environment is inherently difficult for the latter agent. On the other hand, the agent capable of discovering deeper subgoal hierarchies could achieve high performance already after the first round of learning. 

A video demonstrating agent's behavior in the above mentioned environments and a listing of policies learned in the experiments are available at \href{http://shorturl.at/deB24}{shorturl.at/deB24}

\section{Related Work}\label{Sect:RelatedWork} 
In most of the HRL models with subgoal discovery known in the literature the depth of the subgoal hierarchy or the total maximal number of subgoals is restricted by a hyperparameter. 
There are approaches, in which the problem of hierarchical policy learning is solved with subgoal discovery in a unified fashion, and there are proposals focused on the subgoal discovery as the central problem, without any relation to concrete RL models. To the best of our knowledge, there are only two approaches in the first category, in which arbitrarily deep subgoal hierarchies are supported. First, we comment on each of them. 

In \cite{Konidaris2009} a method for skill chaining based on the Options Framework \cite{Sutton1999} is proposed, in which tree-like sequences of options (skills) are built by a backward chaining algorithm, from a goal to subgoals. The initiation states for each option are determined by a state classifier, which discovers those states, from which the achievement of the corresponding (sub)goal state is most likely. The initiation states are then taken as subgoal states and the backward chaining procedure is applied recursively to each of them, thus producing a tree-like option hierarchy. In our model, the policy learning procedure is conceptually similar to the backward skill chaining and each environment rule used in the formulation of a policy could in principle be viewed as a representation of a particular skill (of making a one-step transition between two states). The environment rule learning module could then be seen as an analogue of a state classifier that gives the most probable states, from which a given state is one-step accessible. In our approach however, subgoals differ from skills in that they are learned on top of the previously computed policies and correspond to the descriptions of states, which, if previously achieved, increase the total efficiency of the available policies. Importantly, these state descriptions are introduced as new predicates into the language of the agent, which allows for 
adapting environment rules and policies accordingly.

In \cite{Vezhnevets2017} the ideas of Feudal Reinforcement Learning \cite{FeudalRL} are implemented in a two-level model, in which the higher level (manager) sets subgoals to the lower one (worker) as learned abstractions of important intermediate states on the way to the primary goal. The manager communicates the next subgoal to the worker, but it does not specify how to achieve it. The manager and the worker learn independently of each other and operate at different temporal resolutions, which allows for solving tasks involving long-term credit assignment. This way the model is able to support more abstract (automatically learned) subgoals, e.g., ``obtain X'' in comparison to ``obtain X at location Y''. We note that these features are present in our approach too. Action Planning in our model can be viewed as a procedure, which communicates tasks between the control levels corresponding to the learned subgoal hierarchy (see Figure \ref{Fig:FlowChart}). Abstract tasks (such as, e.g., ``obtain an item of type $i$'') are supported. An important feature of our approach, when formulated in the terms of Feudal RL, is that the model makes a choice of the most appropriate next worker dynamically depending on the current state of the agent and the achieved subgoals.

In most of the approaches to solving the subgoal discovery as a standalone problem subgoals are searched as bottleneck nodes in the State Transition Graph (built for the complete environment or its approximation) by using centrality measures, clustering, or spectral analysis \cite{Mendonca2019}. There are several approaches however, which rely solely on the transition history of an agent instead of the State Transition Graph. For example, \cite{ICML-2001-McGovernB} employs the Options Framework \cite{Sutton1999} and the concept of \emph{diverse density} for discovering bottleneck regions in the agent’s observation space. Agent's trajectories are divided into positive (which bring the agent to a goal) and negative ones. Those states are taken as candidate options, which are often present in positive trajectories and never in negative ones. Initiation states for an option (a subgoal) $G$ are defined as those states, which have been visited some $n$ steps before $G$, where $n$ is a hyperparameter. An additional parameter is used to exclude candidate options, which are too close to the previously discovered subgoals or their initiation states. \cite{Chen2007} employs the same idea for subgoal discovery, but it uses a different filtering mechanism for candidate subgoal states, which leaves only those ones that correspond to local and global visitation maxima. The idea of interpreting subgoals as bottleneck states in the sense above is similar to how we view subgoals in our paper. For subgoal discovery, we analyze trajectories that correspond to the most efficient policies. The difference is that we consider them as an aggregate policy and we compare its efficiency before and after adding the fact of the achievement of a candidate subgoal, which facilitates finding only the most significant subgoals. 

From the point of view of interpretability in RL, in the recent couple of years quite a few models have been proposed in the literature, which are transparent and inherently explainable. Many approaches in interpretable RL are based on a post-hoc analysis and explanation of pretrained black-box models with the help of more simple glass-box models. A slightly different way is followed in \cite{Verma2018}, which employs a pretrained black-box NN model for learning a declarative model, in which policies are given by functional programs. The declarative model is learned to approximate the output of the NN model and it gives human-readable policies, which provide more smooth control in experiments. An important point in this approach is that only those policies are learned that comply with a given syntactic template, the form of which is essentially a hyperparameter dependent on the environment. As in \cite{Verma2018}, the policies in our approach are declarative: they can be viewed as probabilistic rules stating that a sequence of transitions implies the achievement of a (sub)goal with a certain probability. The general form of the rules is not restricted by any domain dependent template, but there are hyperparameters, which allow for fine-tuning the properties of the learned policies. 

In \cite{Hein2017} a rule-based approach is used for the development of fuzzy controllers, which learn via interaction with a NN-simulated environment. Policies are defined as instantiations of fuzzy rule templates, the parameters of which are learned by particle swarm optimization, although evolutionary algorithms or gradient descend are equally applicable. For instance, in the earlier work \cite{Juang2000} evolutionary algorithms have been used for computing both, rule templates and their parameters. It would be useful to apply these algorithms for environment rule and policy learning in our approach and we leave this for future research. Our work however is conceptually different from \cite{Hein2017}, \cite{Juang2000}, and similar approaches in an important aspect. In the named approaches, in each round of learning the complete set of policies is generated in a top-down fashion and it is then evaluated in the environment. In case performance is unsatisfactory, the whole process repeats. In contrast, our approach is based on goal directed policy generation and supports continuous learning on agent's history consisting of random actions and policy guided transitions. Each policy is built upon probabilistic laws, which are environment rules with balanced size and informativeness. This facilitates the readability of the learned policies and contributes to the sample efficiency of our model. 

\section{Discussion and Outlook}\label{Sect:Outlook}
The proposed model uniquely combines the ability to generate human-readable policies and to discover  subgoals, for which an agent does not initially have sensors. Unlike the common RL approaches, the model does not require a reward function to be specified. We believe that the model deserves further development and it can be used as a framework for testing the limits of HRL based on rule learning. Some ideas of hierarchical learning in our model could be employed in the general RL, also for tasks in continuous environments.  

The current version of our model supports only incremental achievement of subgoals: an agent can not lose a previously achieved subgoal (e.g., an item it has picked up) and subgoals can not be conflicting (e.g., picking up an item of one type cannot make picking up an item of another type impossible). In general, the subgoal discovery mechanism should be enhanced in order to support different logical constraints between subgoals imposed by environment. 

In complex scenarios it can be the case that a particular subgoal is rarely accessible (compared to  another subgoal $S$), but once achieved it allows the agent to reach the primary goal faster. The ability to reason about subgoals this way needs to be integrated into the model, because in the current implementation, an agent will tend to prefer $S$ as a subgoal, as most of the trajectories go through $S$. In our model, an agent learns how to act efficiently in situations when a (sub)goal is located in the scope of its sensor field, but the model does not help in exploring the environment in other situations; thus, the exploration problem should further be addressed. 

Some policies learned by the agent may happen to be faulty if it starts learning in some ``atypical'' part of environment. For example, if an agent is initialized in a specific area of our model environment, in which the distribution of items is atypically dense, then it could learn probabilistic laws,
which would not generalize well. If their probability degrades slowly during exploration then these faulty rules will be for a long time preferred for building policies. 
In general, it is important to develop mechanisms that could help the agent to recognize significant changes in the environment in order to recalculate rules and policies accordingly. 

In this paper, we did not discuss the question of optimality of policies. As policies in our model are human-readable, we could confirm in each of our experiments that the agent obtained optimal policies already after the first round of learning. However, this property obviously needs a theoretical investigation. Finally, it is important to note that several components in our model employ  hyperparameters for fine-tuning. 
Much study could be devoted to the influence of these parameters on agent's performance in various environments. However, it would be more interesting to investigate whether the model architecture could live without (at least some of) the hyperparameters, which is also a part of our further research.

\bibliographystyle{kr}
\bibliography{hrl-subgoal}

\end{document}